\pdfoutput=1
\documentclass[11pt]{article}

\usepackage[preprint]{acl}

\usepackage{times}
\usepackage{latexsym}

\usepackage[T1]{fontenc}
\usepackage[utf8]{inputenc}

\usepackage{microtype}

\usepackage{inconsolata}

\usepackage{graphicx}

\usepackage{amsmath,amssymb}
\usepackage{hyperref}
\usepackage{multicol}
\usepackage{multirow}
\usepackage{setspace}
\usepackage{multicol}
\usepackage{rotating}
\usepackage{booktabs}
\usepackage{makecell}
\usepackage{xcolor}
\usepackage{array}
\usepackage{subcaption}

\title{Are Expert-Level Language Models Expert-Level Annotators?}

\author{Yu-Min Tseng$^\alpha$$^\beta$\quad 
        Wei-Lin Chen$^\alpha$$^\gamma$\quad
        Chung-Chi Chen$^\delta$\quad
        Hsin-Hsi Chen$^\alpha$
    \vspace{5pt} \\
    $^\alpha$National Taiwan University\quad
    $^\beta$Academia Sinica\quad
    $^\gamma$University of Virginia\quad
    $^\delta$AIST, Japan
    \vspace{2pt}\\
    \texttt{ymtseng@nlg.csie.ntu.edu.tw\quad wlchen@virginia.edu}\\
    \texttt{c.c.chen@acm.org\quad hhchen@ntu.edu.tw}
}

\begin{document}

\maketitle

\begin{abstract}
Data annotation refers to the labeling or tagging of textual data with relevant information.
A large body of works have reported positive results on leveraging LLMs as an alternative to human annotators.
However, existing studies focus on classic NLP tasks, and the extent to which LLMs as data annotators perform in domains requiring expert knowledge remains underexplored.
In this work, we investigate comprehensive approaches across three highly specialized domains and discuss practical suggestions from a cost-effectiveness perspective.
To the best of our knowledge, we present the first systematic evaluation of LLMs as expert-level data annotators.
\end{abstract}

\section{Introduction}
Data annotation refers to the task of labeling or tagging textual data with relevant information~\cite{tan2024large}.
For example, adding topic keywords to social media contents.
Typically, data annotation is carried out by crowd-sourced workers (\textit{e.g.}, MTurkers) or specialized annotators (\textit{e.g.}, researchers), depending on the tasks, to ensure high-quality annotations.
However, the annotating procedures are often costly, time-consuming, and labor-intensive, particularly for tasks that require domain expertise.

With the rise of large language models (LLMs), a series of works have explored using them as an attractive alternative to human annotators~\cite{ding2023gpt,zhang2023llmaaa,choi2024gpts,he2023annollm}.
Empirical results show that, in certain scenarios, LLMs such as ChatGPT and GPT-3.5 even outperform master-level MTurk workers, with substantially lower per-annotation cost~\cite{gilardi2023chatgpt,alizadeh2023open,bansal2023large,zhu2023can}.
However, existing studies mainly focus on classic NLP tasks (\textit{e.g.}, sentiment classification, word-sense disambiguation) on general domain datasets.
The extent to which LLMs as data annotators perform in domains requiring expert knowledge remains unexplored.
\begin{figure}[t]
  \centering
  \includegraphics[width=0.9\linewidth]{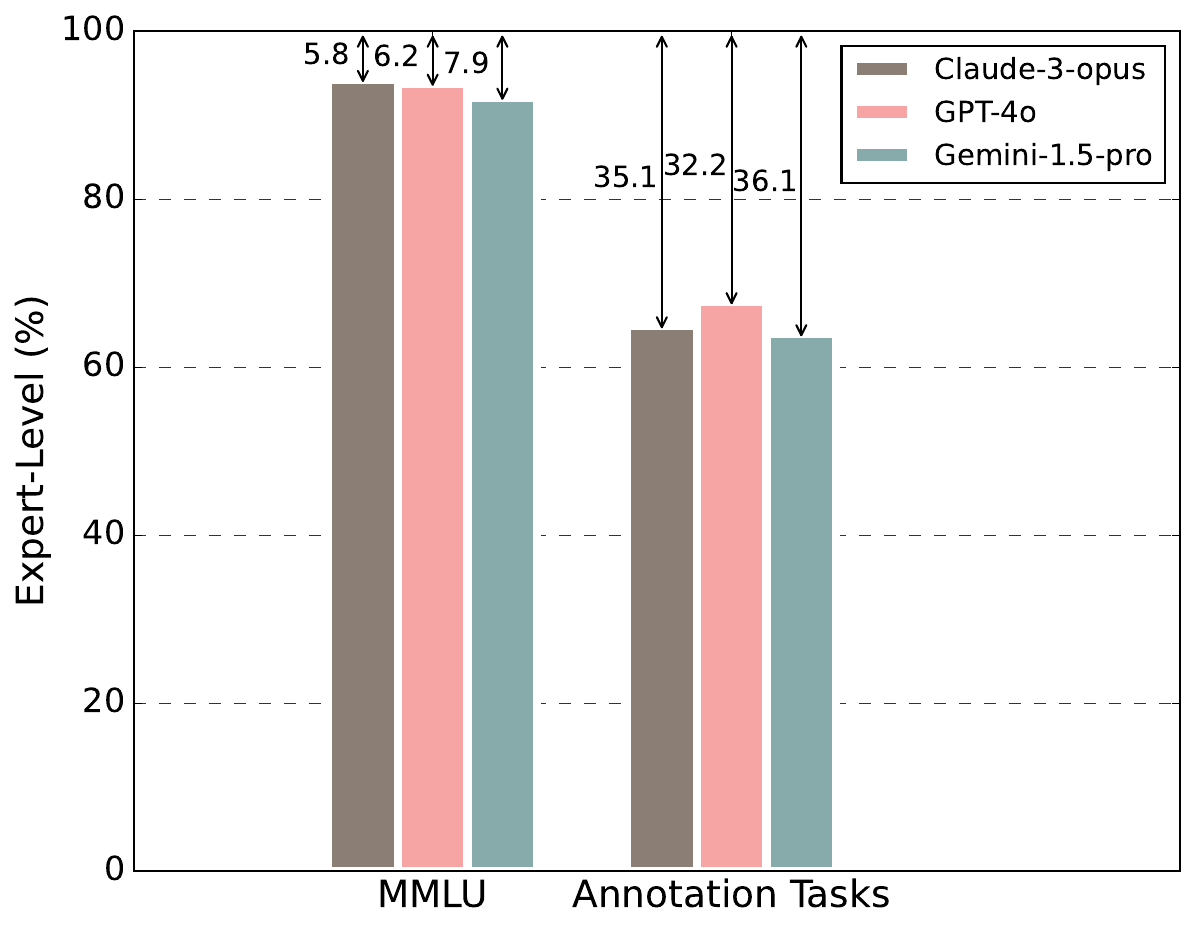}
  \caption{The degree of expert-level performance reached by state-of-the-art (SOTA) LLMs. For MMLU, we report model scores from the HELM~\cite{liang2023holistic} website divided by human-expert score (89.8) from~\citet{hendrycks2020measuring}.}
  \label{fig:illustration}
\end{figure}

On the other hand, LLMs have exhibited striking performance in a variety of benchmarks, both professional and academic~\cite{jin2019pubmedqa,hendrycks2020measuring,chen2021evaluating,rein2023gpqa,achiam2023gpt}.
Leveraging the abundant domain-specific knowledge encoded in the parameters, LLMs could pass exams that require expert-level abilities~\cite{choi2021chatgpt,singhal2023large,callanan2023can,singhal2023towards,katz2024gpt}.
These findings prompt the question -- Can LLMs apply their parametric knowledge to perform expert-level annotation tasks?

To address this, we investigate three specialized domains: finance, biomedicine, and law.
Specifically, we adopt six existing datasets that
\textit{(i)} provide fully-detailed annotation guidelines and
\textit{(ii)} are manually labelled by domain experts.
We format the annotation task, the guideline, and unlabelled data instances as instructional inputs to the most performant, publicly-available LLMs, and evaluate their annotation results against ground-truth labelled by human experts.
Experimental results in our vanilla setting suggest that LLMs show substantial rooms for improvements, with an average of around 35\% behind human expert annotators.

Towards a more comprehensive evaluation, we employ a variety of approaches tailored to elicit the capabilities in LLMs, including chain-of-thought (CoT), self-consistency, and self-refine promptings.
Additionally, drawing inspiration from how human annotators reach consensus, we introduce a multi-agent annotation framework which incorporates a peer-discussion process for producing annotations.
Lastly, we discuss practical suggestions on leveraging LLMs for expert annotation tasks, from a cost-effectiveness perspective.
We summarized our main contributions as follows:
\begin{itemize}
    \item We present, to the best of our knowledge, the first systematic evaluation of LLMs as expert-level data annotators.
    \item We explore comprehensive approaches, including prompt-based methods and multi-agent frameworks, across three highly specialized domains.
    \item We provide a cost-effectiveness analysis and practical suggestions on leveraging LLMs for expert annotation tasks.
\end{itemize}

\begin{table*}[htbp]
  \small
  \centering
  \resizebox{\textwidth}{!}{
    \begin{tabular}{lrrrrrrrrrr}
    \toprule
    \multirow{2}[4]{*}{\textbf{Model / Method}} & \multicolumn{2}{c}{\textbf{Finance}} &       & \multicolumn{2}{c}{\textbf{Biomedicine}} &       & \multicolumn{2}{c}{\textbf{Law}} &       & \multirow{2}[4]{*}{\textbf{Avg.}} \\
\cmidrule{2-3}\cmidrule{5-6}\cmidrule{8-9}          & \multicolumn{1}{c}{\textbf{REFinD}} & \multicolumn{1}{c}{\textbf{FOMC}} &       & \multicolumn{1}{c}{\textbf{AP-Rel}} & \multicolumn{1}{c}{\textbf{CODA-19}} &       & \multicolumn{1}{c}{\textbf{CUAD}} & \multicolumn{1}{c}{\textbf{FoDS}} &       &  \\
    \midrule
    GPT-3.5-Turbo & 47.4  & 60.4  &       & 58.9  & 64.4  &       & 71.8  & 37.1  &       & 56.7 \\
    GPT-4o & \textbf{67.2} & \textbf{67.6} &       & 65.8  & \textbf{79.3} &       & \textbf{82.2} & 44.4  &       & \textbf{67.8} \\
    Gemini-1.5-Pro & 64.6  & \textbf{67.6} &       & 54.8  & 73.2  &       & 80.6  & 42.8  &       & 63.9 \\
    Claude-3-Opus & 61.2  & 63.6  &       & \textbf{71.2} & 65.6  &       & 80.8  & \textbf{46.9} &       & 64.9 \\
    \midrule
    \textit{GPT-4o} & 67.2  & 67.6  &       & 65.8  & 79.3  &       & 82.2  & 44.4  &       & 67.8 \\
    \,\,\,\,CoT   & 71.0 (\textcolor{teal}{$\uparrow$3.8}) & $^*$68.2 (\textcolor{teal}{$\uparrow$0.6}) &       & $^*$68.5 (\textcolor{teal}{$\uparrow$2.7}) & $^*$81.1 (\textcolor{teal}{$\uparrow$1.8}) &       & 79.8 (\textcolor{purple}{$\downarrow$2.4}) & 43.9 (\textcolor{purple}{$\downarrow$0.5}) &       & 68.7 \\
    \,\,\,\,Self-Consistency & $^*$72.4 (\textcolor{teal}{$\uparrow$5.2}) & $^*$70.4 (\textcolor{teal}{$\uparrow$2.8}) &       & $^*$68.5 (\textcolor{teal}{$\uparrow$2.7}) & 78.9 (\textcolor{purple}{$\downarrow$0.4}) &       & $^{*\dagger}$82.4 (\textcolor{teal}{$\uparrow$0.2}) & $^{*\dagger}$45.0 (\textcolor{teal}{$\uparrow$0.6}) &       & \textbf{69.6} \\
    \,\,\,\,Self-Refine & 70.0 (\textcolor{teal}{$\uparrow$2.8}) & $^*$69.2 (\textcolor{teal}{$\uparrow$1.6}) &       & $^*$69.9 (\textcolor{teal}{$\uparrow$4.1}) & $^*$81.5 (\textcolor{teal}{$\uparrow$2.2}) &       & 78.0 (\textcolor{purple}{$\downarrow$4.2}) & $^*$45.5 (\textcolor{teal}{$\uparrow$1.1}) &       & 69.0 \\
    \bottomrule
    \end{tabular}%
    }
  \caption{The performance of SOTA LLMs as annotators (accuracy) and a comparison of GPT-4o with different advanced techniques for expert-level annotation tasks. An asterisk ($^*$) indicates that the method is statistically significant with p-value < 0.05 than the vanilla method. A dagger ($^\dagger$) indicates that the self-consistency method is statistically significant with p-value < 0.05 than the CoT method.}
  \label{tab:main}%
\end{table*}%

\section{Datasets}
\paragraph{Finance}
We adopt the REFinD~\citep{kaur2023refind} and FOMC datasets~\citep{shah2023trillion} for financial domain.
REFinD is the largest relation extraction dataset over financial documents, comprising 8 entity pairs and 29 relations, with labels reviewed by financial experts.
In this task, annotators are tasked to extract relations between finance-specific entity pairs, such as [\textit{person}] is an employee of [\textit{organization}].
FOMC is constructed for identifying sentiments about the future monetary policy stances, annotated by experts with a correlated financial knowledge.
The labels of this annotation task are Dovish, Hawkish, and Neutral, where a Dovish sentence indicates easing and a Hawkish sentence indicates tightening.

\paragraph{Biomedicine}
For the biomedical domain, we utilize AP-Relation dataset~\citep{gao2022hierarchical} and COVID-19 Research Aspect Dataset (CODA-19;~\citealp{huang2020coda}).
AP-Relation is designed for extracting the relationship between Assessment and Plan Subsections in daily progress notes.
The Assessment describes the patient and establishes the main symptoms or problems for their encounter, while the Plan Subsection addresses each differential diagnosis or problem with a daily action or treatment plan.
The annotation label schemes for different relations are categorized as \textit{direct}, \textit{indirect}, \textit{neither}, or \textit{not relevant}.
CODA-19 codes each segment aspect of English abstracts in the COVID-19 Open Research Dataset~\citep{wang2020cord}.
In this task, annotators are tasked to label each segment as \textit{background}, \textit{purpose}, \textit{method}, \textit{finding/contribution}, or \textit{other} sections.
To ensure the quality of the labels, we only adopt instances annotated by biomedical experts.

\paragraph{Law}
In the legal domain, we adopt Contract Understanding Atticus Dataset (CUAD;~\citealp{hendrycks2021cuad}) and Function of Decision Section dataset (FoDS;~\citealp{guha2024legalbench}).
CUAD consists of legal contracts with extensive annotations from legal experts, created with a year-long effort by dozens of law student annotators, lawyers, and machine learning researchers.
The annotation task is to label 41 types out of legal clauses, classified into 5 answer categories, that are considered important in contract review related to corporate transactions.
We manually use ``Yes/No'' answer category to construct our annotation task as the identification of 32 types of clauses.
FoDS comprises one-paragraph excerpts from legal decisions, annotated by legal professionals who are included as authors.
In this task, annotators are tasked to review a legal decision and identify one out of seven function categories that each section (\textit{i.e.}, excerpt) of the decision serves.
We provide statistics and annotation guidelines of each dataset in~\ref{subsec:appendix-statistics} and~\ref{subsec:appendix-guideline}.

\section{LLMs as Expert Annotators}\label{sec:llm-as-anno}
\subsection{Methods}
\paragraph{Vanilla}
The vanilla method refers to standard direct-answer prompting, where instructional input consists of the annotation task, guideline, and the sample to be annotated are given to the LLMs.
LLMs are tasked to conduct annotation as a domain expert of relevant fields.
We utilized a uniform prompt template that is easily generalizable across domains and datasets.
The vanilla prompt also serves as the base of other sophisticated approaches (described below).
We provide all prompt templates in~\ref{subsec:appendix-prompt}.

\paragraph{CoT} Prompting with chain-of-thought (CoT) improves LLMs' complex reasoning ability significantly~\cite{wei2022chain}.
Specifically, we employ zero-shot CoT~\cite{kojima2022large}, where a trigger phrase ``\textit{Let’s think step by
step}'' augments the prompt to elicit reasoning chain from LLMs and leads to a more accurate answer.

\paragraph{Self-Consistency}
Self-consistency~\cite{wang2022self} further improves upon CoT via a sample-and-marginalize decoding procedure, which selects the most consistent answer rather than the greedily decoded one.
Concretely, we sample 5 diverse reasoning paths with temperature 0.7, and take the majority vote to determine the final answer.

\paragraph{Self-Refine}
The self-refine~\cite{madaan2024self} method includes three steps: generate, review, and refine.
An LLM first generates an initial answer (\textit{i.e.}, draft).
Then, the model review its draft and provide feedback.
Lastly, the LLM refine the draft by incorporating  its feedback, and outputs an improved answer.
The same LLM is used in all steps.

\subsection{Results}
We report our main results in Table~\ref{tab:main}.
We compare four models, including GPT-3.5-Turbo~\cite{openai2023gpt}, GPT-4o~\cite{openai2024gpt4o}, Gemini-1.5-Pro~\cite{reid2024gemini}, and Claude-3-Opus~\cite{anthropic2024claude}, and report their annotation accuracy.
We use labels annotated by human experts from the corresponding dataset as ground-truth answers.

As observed, under the vanilla method (upper block), GPT-4o records the best overall performance.
Claude-3-Opus and Gemini-1.5-Pro achieve similar scores, while GPT-3.5-Turbo performs notably worse.
However, all LLMs show substantial rooms for improvements, with an average of 32.2\% $\sim$ 43.3\% behind human expert annotations.
The best single score (GPT-4o on CUAD dataset) still lacks around 20\%.
The results suggest that naive standard prompting is \textit{not} feasible to obtain satisfactory annotation quality from LLMs in tasks involving domain expertise.

To probe the capabilities of LLMs more further, we experiment GPT-4o with three methods: CoT, self-consistency (SC), and self-refine (SR), proposed to improve LLMs factual knowledge and reasoning capabilities.
The results are present in~Table~\ref{tab:main} lower block.
As observed, in general, all methods exhibit improved results, with an average of 1\% $\sim$ 2\% accuracy gain.
However, comparing with the huge performance boosts of how these methods typically benefit general domain datasets, their efficacy on expert-level annotation tasks is relatively low.
This might imply that the models inherently lack necessary knowledge and reasoning capability to perform as expert annotators.

\begin{figure}[t]
  \centering
  \includegraphics[width=0.95\linewidth]{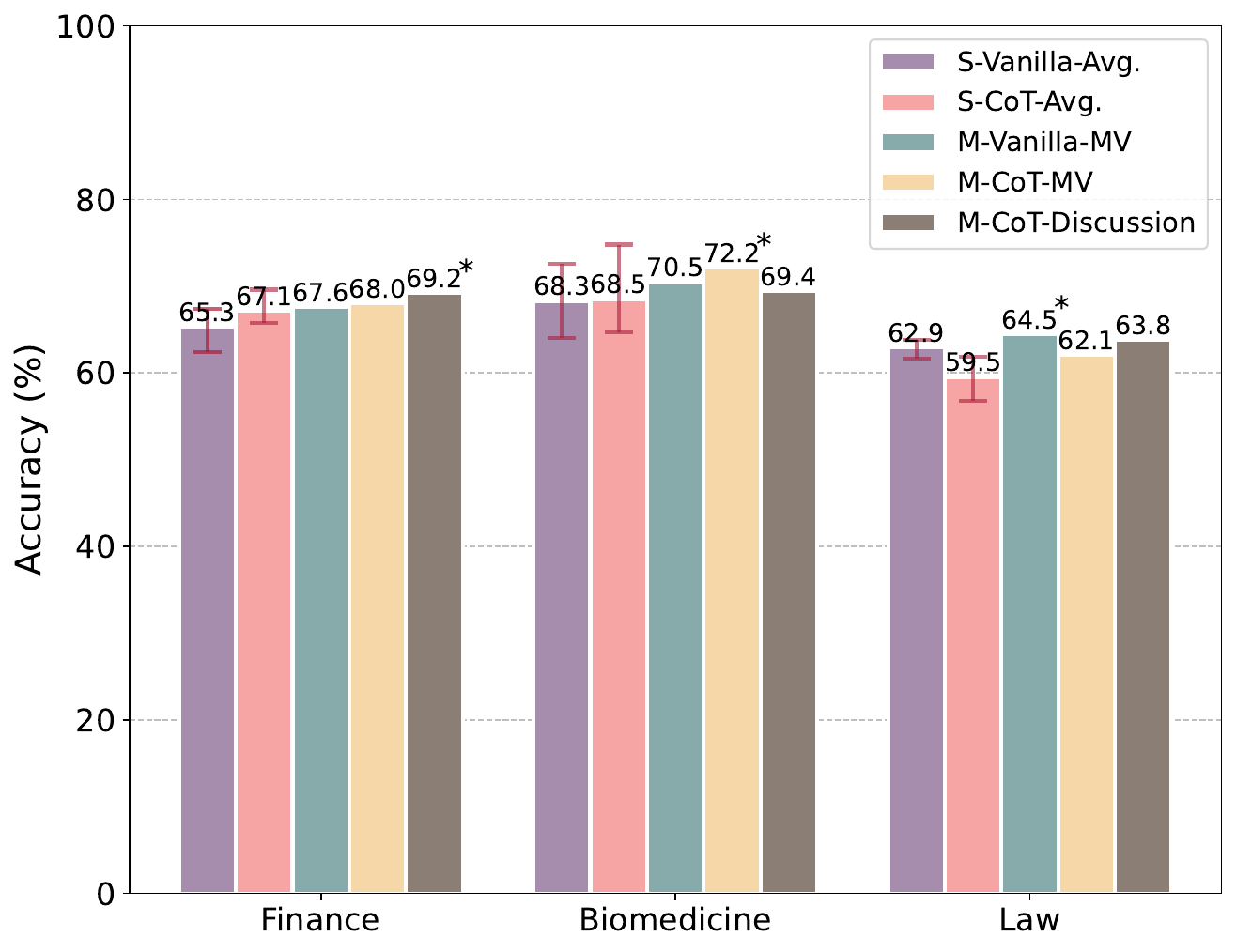}
  \caption{The performance comparison of different single LLM settings ($S$) and multi-agent frameworks ($M$) across three domains. For the two single agent settings, numbers on the figure represent the average performance of the three single LLMs: GPT-4o, Gemini-1.5-Pro, and Claude-3-Opus, and red bars indicate the range of performance. An asterisk ($^*$) indicates that the method is statistically significant with p-value < 0.05.}
  \label{fig:single-multi}
\end{figure}

\begin{figure}[t]
  \centering
  \includegraphics[width=0.905\linewidth]{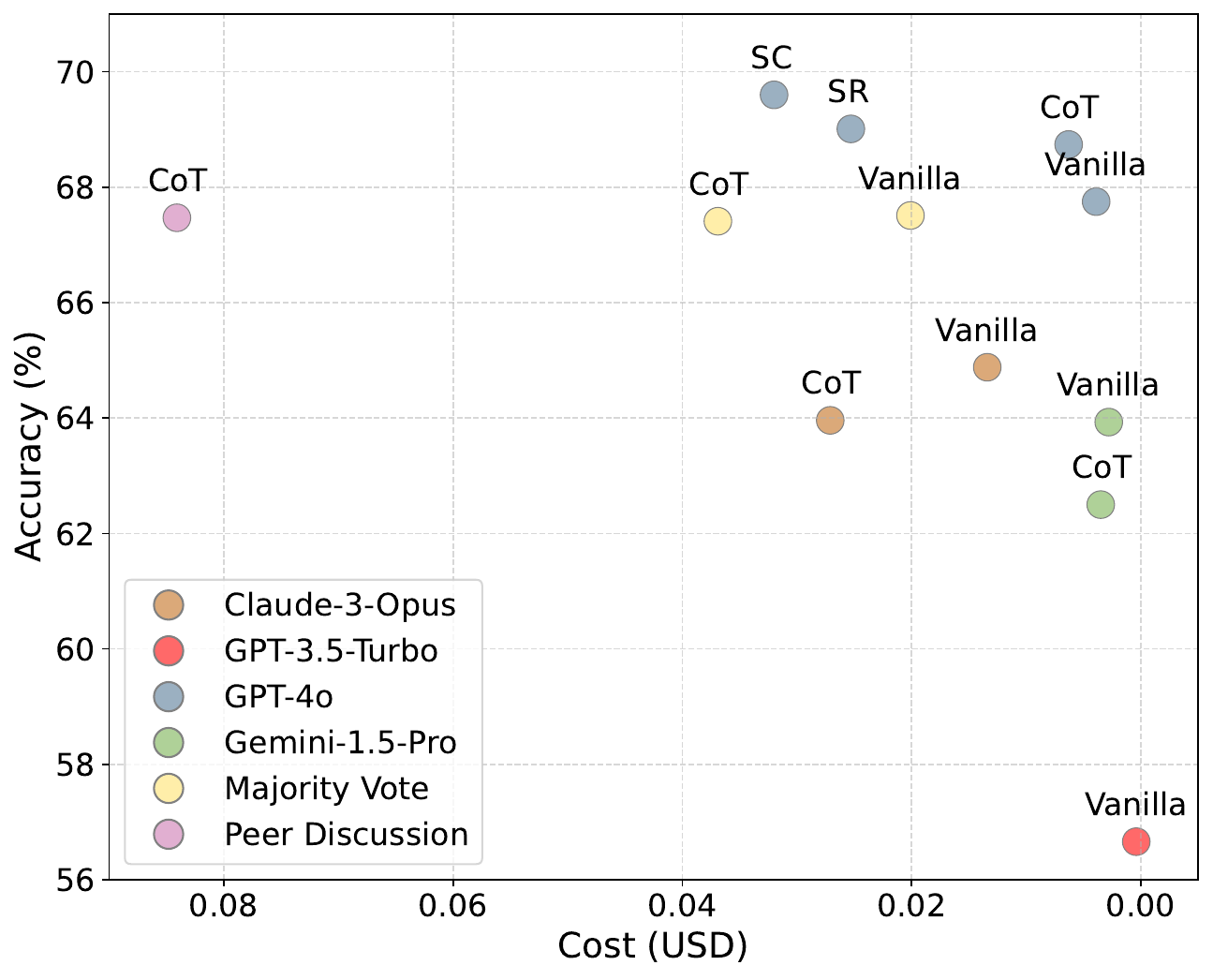}
  \caption{An illustration of the cost-effectiveness relationship of various setups. The x-axis represents the cost per instance in USD, and the y-axis represents the accuracy in percentage. Note that the x-axis is counter-intuitive compared to the usual orientation, with higher costs on the left and lower costs on the right. The upper right corner of the figure indicates better performance, combining lower cost and higher accuracy.}
  \label{fig:cost-acc}
\end{figure}

\section{Multi-Agent Annotation}
The multi-agent framework, where multiple language agents communicate with each other to solve tasks in a collaborative manner, has become a prevalent research direction~\cite{liang2023encouraging,du2023improving,chen2023reconcile,tseng2024two}.
A common scenario in annotation is the disagreement among multiple annotators.
A typical way for resolving such discrepancy is by discussing with others to reach a consent.

Motivated by this, we design a multi-agent annotation framework, which incorporates a peer-discussion process mimicking human annotators for better annotations.
Our multi-agent annotation framework consists of three LLMs: GPT-4o, Gemini-1.5-pro, and Claude-3-opus.

\subsection{Methods}
\paragraph{Majority Vote}
Majority vote (MV) represents a minimal form of discussion, reducing the process to simply selecting the majority output as the final annotation.
We apply two settings for MV: vanilla and CoT.

\paragraph{Peer-Discussion}
Peer-Discussion consists of three steps: (\textit{1}) Generate initial annotation, (\textit{2}) Check annotations, (\textit{3}) Discuss and re-annotate.
Initially, each agent generates their own annotation through CoT prompting given the same annotation task, guideline, and instance.
Next, we check if consensus has been reached (\textit{i.e.}, all annotations are the same labels).
If consensus is achieved, the instance is successfully annotated and the annotation process is complete.
Otherwise, we incorporate all agents' reasoning and labels to generate a ``\textit{Discussion History}''.
Subsequently, agents are required to re-annotate the instance, given the same input and the discussion history.
Thus, we iteratively repeat the same check-consensus-discuss-re-annotate procedure until achieving consensus or reach the maximum discussion round.
In our experimental settings, we set the maximum discussion round to 2.
We provide the prompt templates in~\ref{subsec:appendix-prompt}.

\begin{figure*}[t!]
  \centering
  \includegraphics[width=0.9\linewidth]{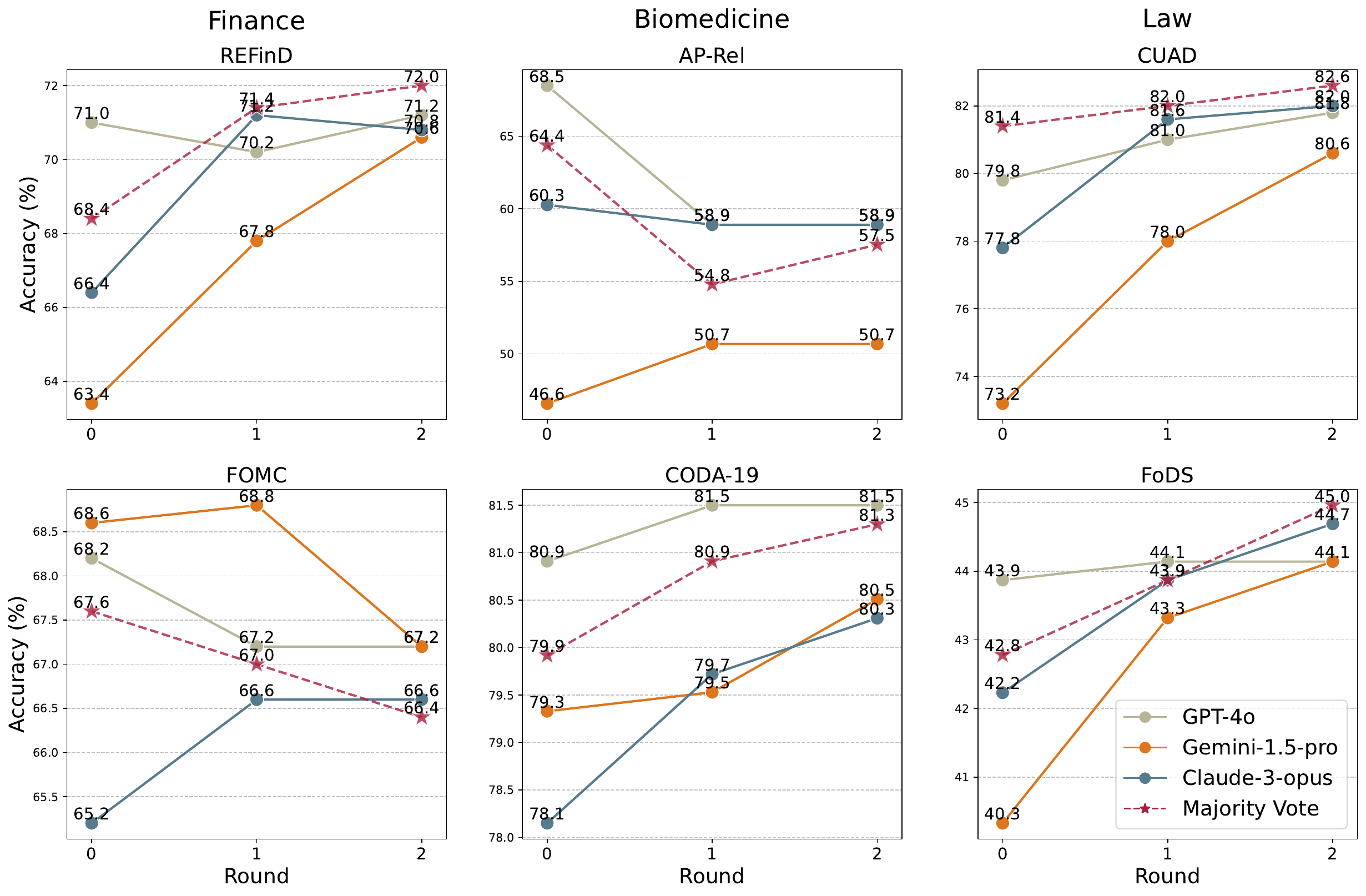}
  \caption{Marginal performance of each LLM during multi-agent peer-discussion process. The performance in Round 0 indicates the LLMs' initial annotation performance, while the performance in Round 1 and Round 2 indicates the LLMs' annotation performance after one or two rounds discussion, respectively.}
  \label{fig:discussion-margin}
\end{figure*}

\subsection{Results}
We present results of multi-agent framework on six datasets with three domains in Figure~\ref{fig:single-multi}, along with results of vanilla and CoT single-LLM settings.

As shown, multi-agent frameworks slightly outperform the average of single-LLM settings.
However, the most performant singe LLM (primarily GPT-4o) still exhibits superior results in both vanilla and CoT setting, similar to the observation by~\citet{wang2024rethinking}.
Though multi-agent methods are still inferior to the best single-LLM setting, it may be a more suitable approach when we are unable to infer which model to adopt in advance.

For each domain, different multi-agent settings achieve the highest performance.
These results demonstrate that the effectiveness of various methods varies across models and datasets, indicating that their utility might be context-dependent.
Additionally, it is noteworthy that the performance of all settings still lag significantly behind human domain-expert annotators.

To further investigate the discussion process in the multi-agent peer-discussion setting, we present the marginal performance of each LLM over two rounds in Figure~\ref{fig:discussion-margin}.
As observed, the peer-discussion process benefits most datasets, leading to improved annotation performance (i.e., increasing \textit{Majority Vote} lines in the figure).
For the AP-Rel dataset, however, peer discussion alternately hurt the performance.
We hypothesize a potential reason is that AP-Rel contains many highly specialized, biomedical-domain specific entities, which are especially challenging for the models.

Experimental results suggest that GPT-4o tends to provide good insight to others, in the mean time, maintaining its own correct answer, especially shown in the REFinD, CUAD, and FoDS datasets.
Under the circumstances, not only does the multi-agent framework performance increase, the individual performance of the three LLMs also benefits from the discussion process.
In contrast, Gemini-1.5-Pro exhibit a tendency to follow and agree with others points of views, instead of defending its own correct annotation, notably demonstrated within the FOMC and CODA-19 datasets.
Therefore, affecting the final discussion performance and its own performance.

\section{Cost-Effectiveness Discussion}
We aggregate our empirical results and compile a cost-effectiveness illustration in Figure~\ref{fig:cost-acc}.
In sum, GPT-4o with vanilla or CoT method presents as the best cost-effective options within our experiments.
GPT-4o with SC achieves the best overall performance at the expense of tripling the cost.
Considering to the real-world annotation scenarios, there are no oracles to determine that which LLM could perform the most accurately against a required annotation task.
Thus, an intermediate option would be multi-agent frameworks, which demonstrate competitive performance and could be a more robust option when access to different LLMs are available.
Despite LLMs do not present as a direct alternative for annotation tasks requiring domain expertise, their collective performance of over 50\% and profoundly lower cost present a promising human-LLM hybrid annotation schema in the future.

\section{Conclusion}\label{sec:discussion}
In this work, we present a pilot focused contribution, as a first attempt, towards answering the research question -- “Can top-performing LLMs, which might be perceived as having expert-level proficiency in academic and professional benchmarks, be an alternative of human-expert annotators?”.
The core finding, from the empirical results on both single-LLM and multi-agent methods across three domains with six datasets, indicates that the current generation of LLMs do \textit{not} present as a direct alternative for annotation tasks requiring domain expertise.
Solely relying on the parametric knowledge in LLMs to perform domain-specific, expert-level annotation tasks is non-trivial.
Considering that these specialized domains are often relevant to high-risk sectors (\textit{e.g.}, medical application), it is crucial to ensure the annotated data has a higher precision and accuracy.

\section*{Limitations}
As we aim to provide direct insight and observation on whether top-performing LLMs can perform as expert annotators \textit{out-of-the-box}, we minimize efforts in prompt engineering.
Some works have demonstrated that, for specific scenarios, one can achieve sizable improvement through carefully-crafted prompts.
Consequently, our results may further benefit from a more exhaustive prompt optimization.

Another potential limitation is that we primarily focus on natural language understanding (NLU) tasks with fixed label space.
Towards a more comprehensive evaluation, natural language generation (NLG) tasks could be further incorporated.

% \section*{Acknowledgements}

% Bibliography entries for the entire Anthology, followed by custom entries
%\bibliography{anthology,custom}
% Custom bibliography entries only
\bibliography{custom}

\appendix

\section{Implementation Details}\label{sec:appendix-implementation}

\subsection{Dataset Statistics}\label{subsec:appendix-statistics}
We provide data statistics of the six existing specialized datasets in Table~\ref{tab:dataset-statistic}.

\subsection{Annotation Guidelines}\label{subsec:appendix-guideline}
\begin{figure*}[t]
  \centering
  \includegraphics[width=\linewidth]{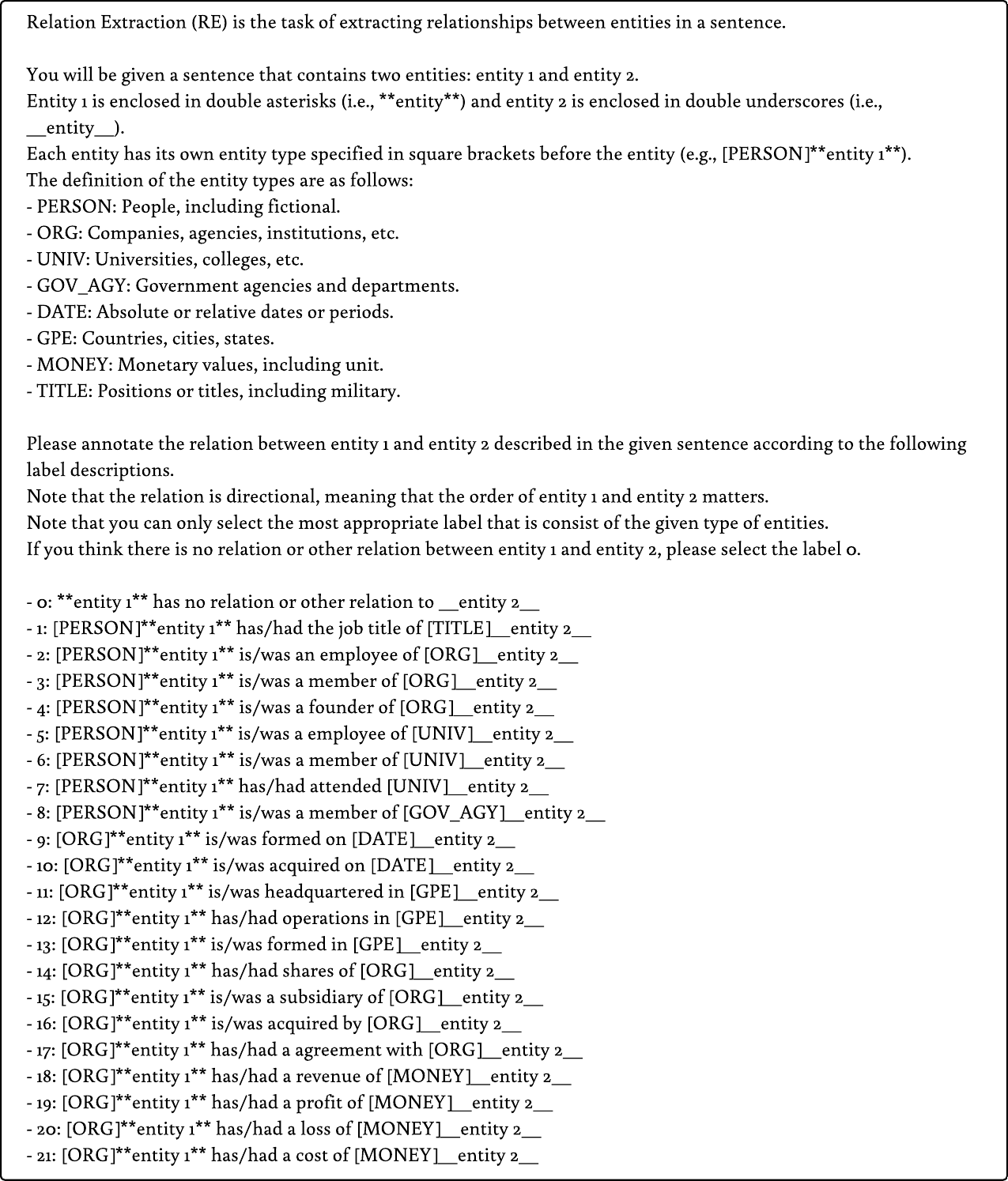}
  \caption{The annotation guideline of REFinD dataset.}
  \label{fig:guideline-refind}
\end{figure*}

\begin{figure*}[t]
  \centering
  \includegraphics[width=\linewidth]{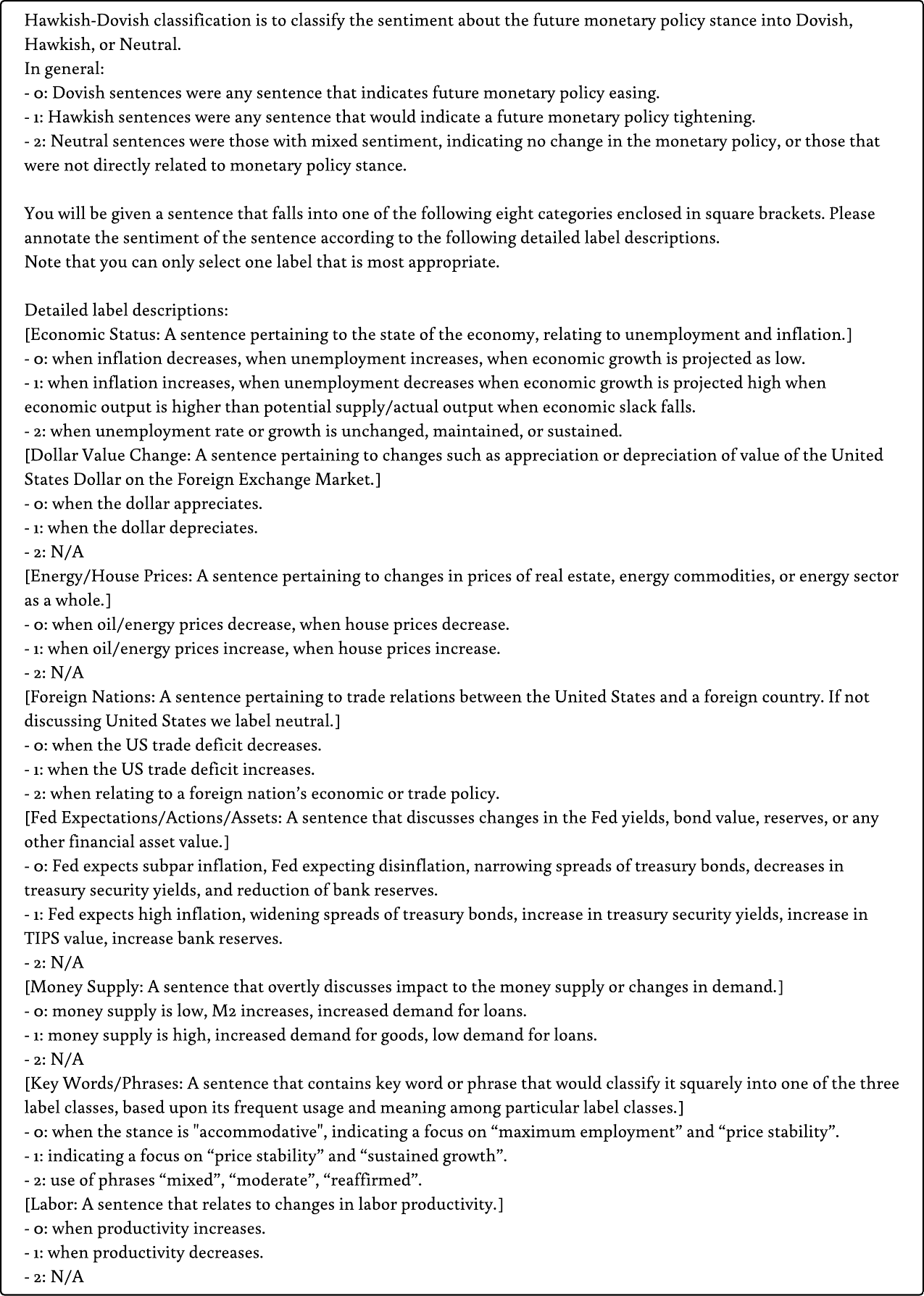}
  \caption{The annotation guideline of FOMC dataset.}
  \label{fig:guideline-fomc}
\end{figure*}

\begin{figure*}[t]
  \centering
  \includegraphics[width=\linewidth]{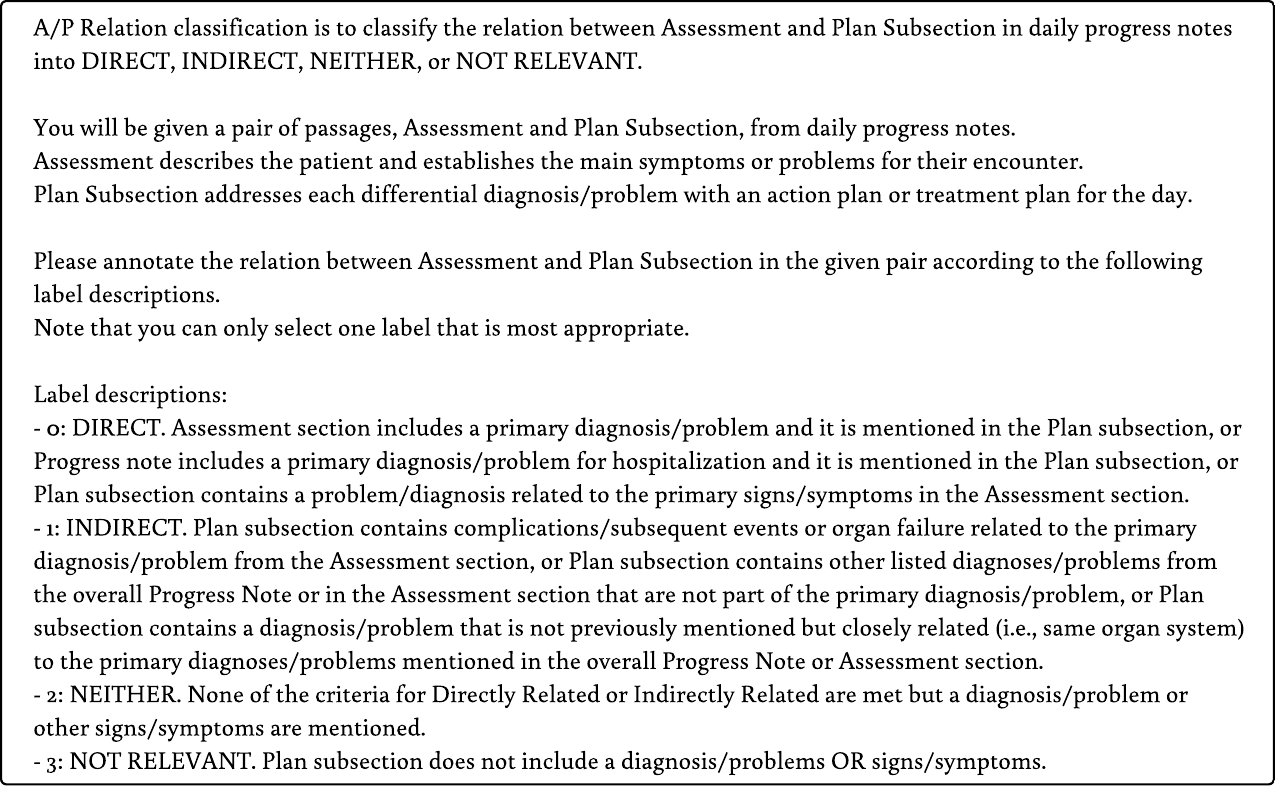}
  \caption{The annotation guideline of AP-Relation dataset.}
  \label{fig:guideline-ap-relation}
\end{figure*}

\begin{figure*}[t]
  \centering
  \includegraphics[width=\linewidth]{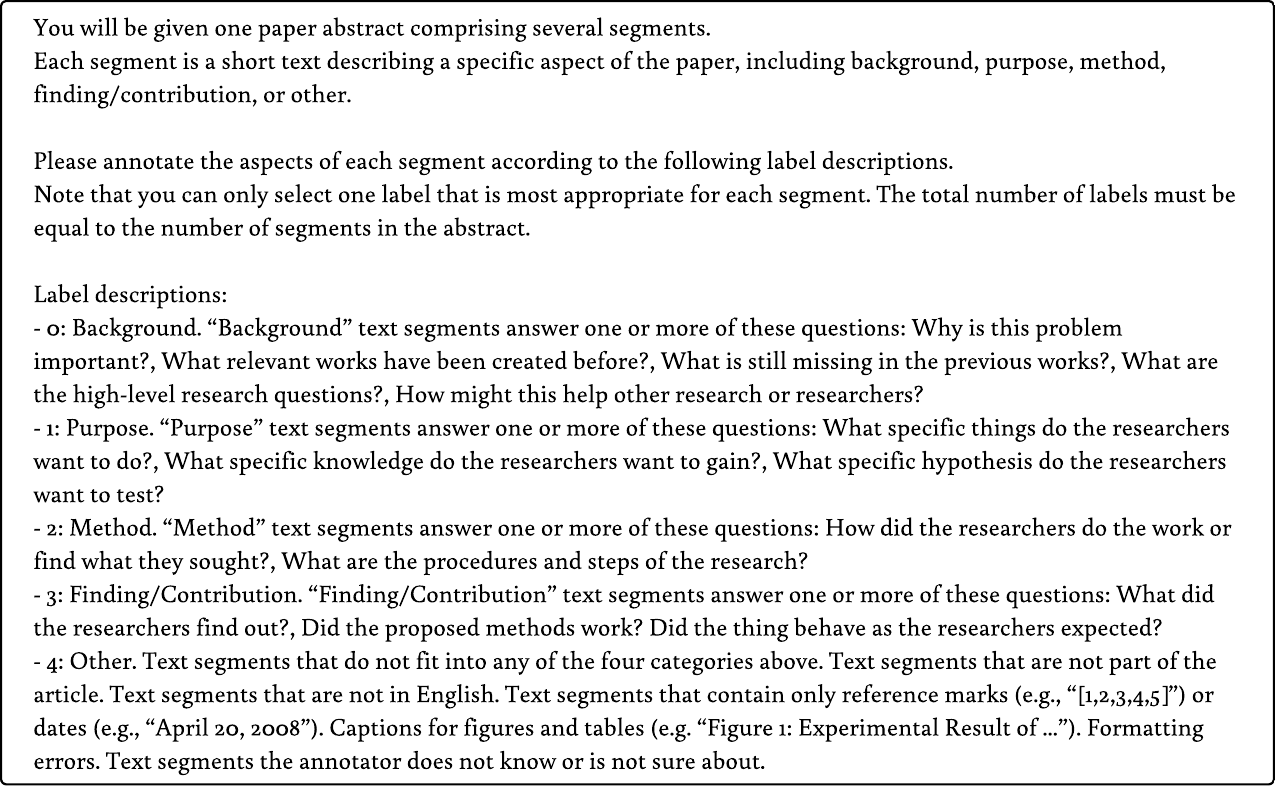}
  \caption{The annotation guideline of CODA-19 dataset.}
  \label{fig:guideline-coda-19}
\end{figure*}

\begin{figure*}[t]
  \centering
  \includegraphics[width=\linewidth]{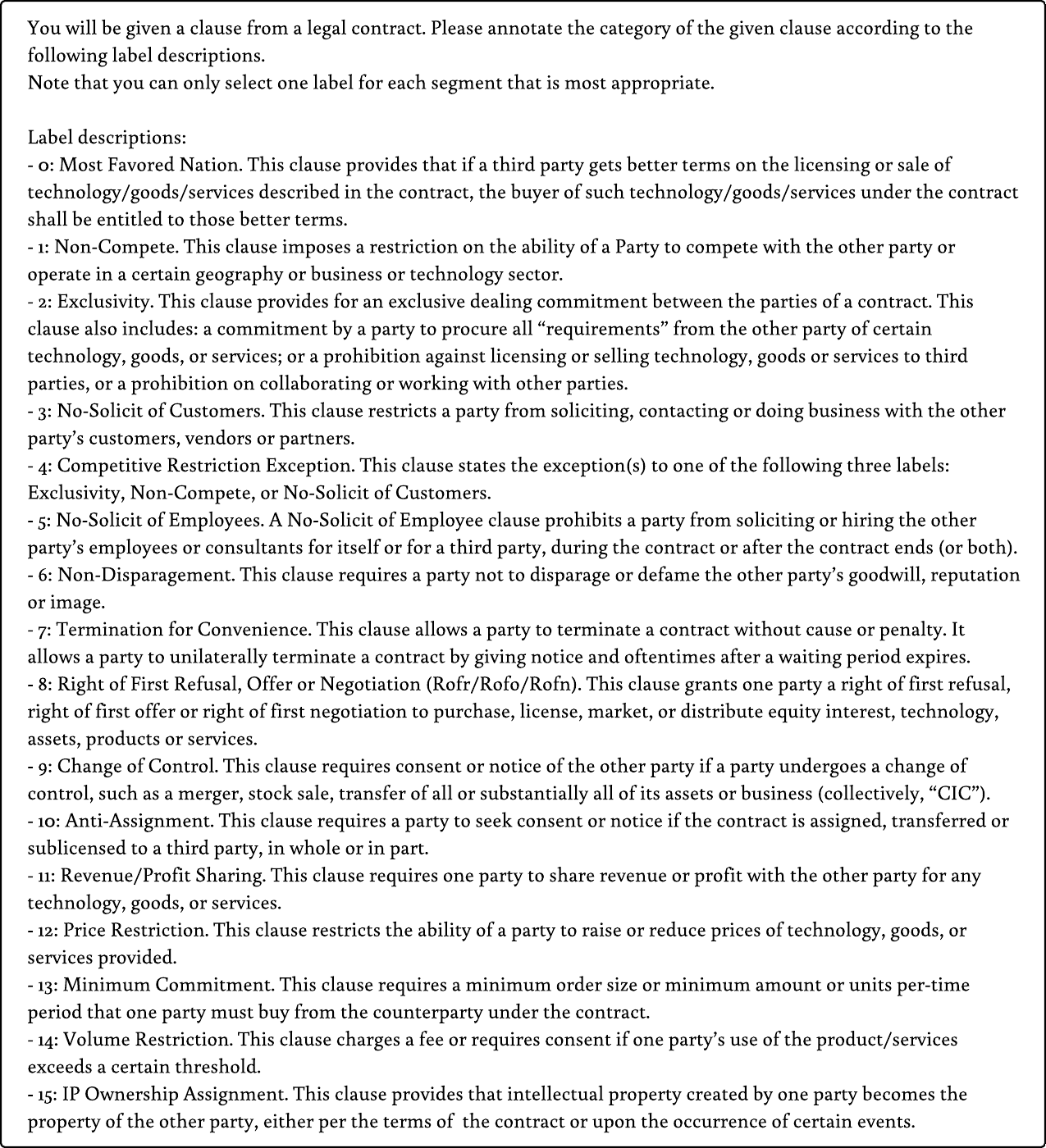}
  \caption{The annotation guideline of CUAD dataset (1-1).}
  \label{fig:guideline-coda-19-1}
\end{figure*}

\begin{figure*}[t]
  \centering
  \includegraphics[width=\linewidth]{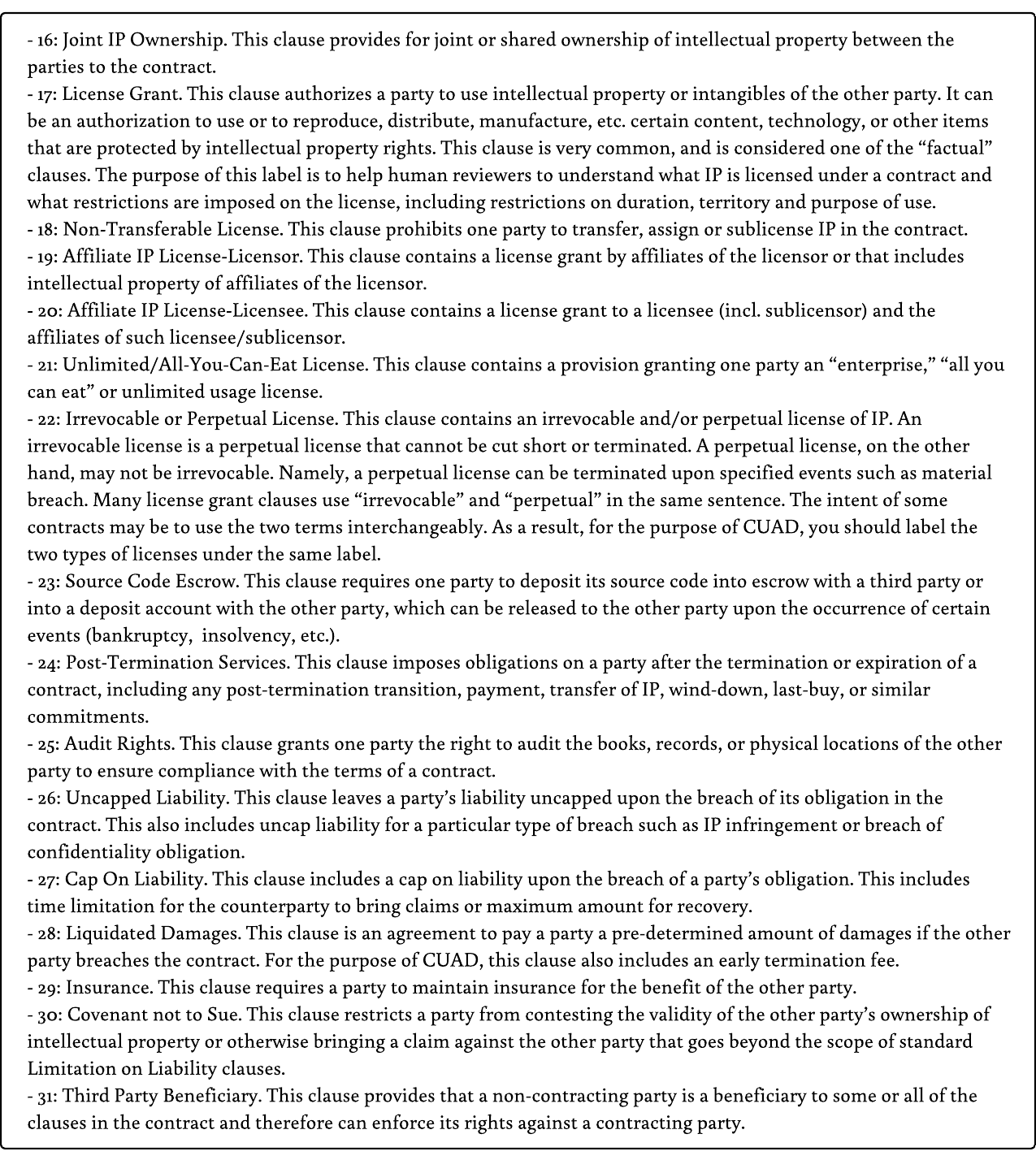}
  \caption{The annotation guideline of CUAD dataset (1-2).}
  \label{fig:guideline-coda-19-2}
\end{figure*}

\begin{figure*}[t]
  \centering
  \includegraphics[width=\linewidth]{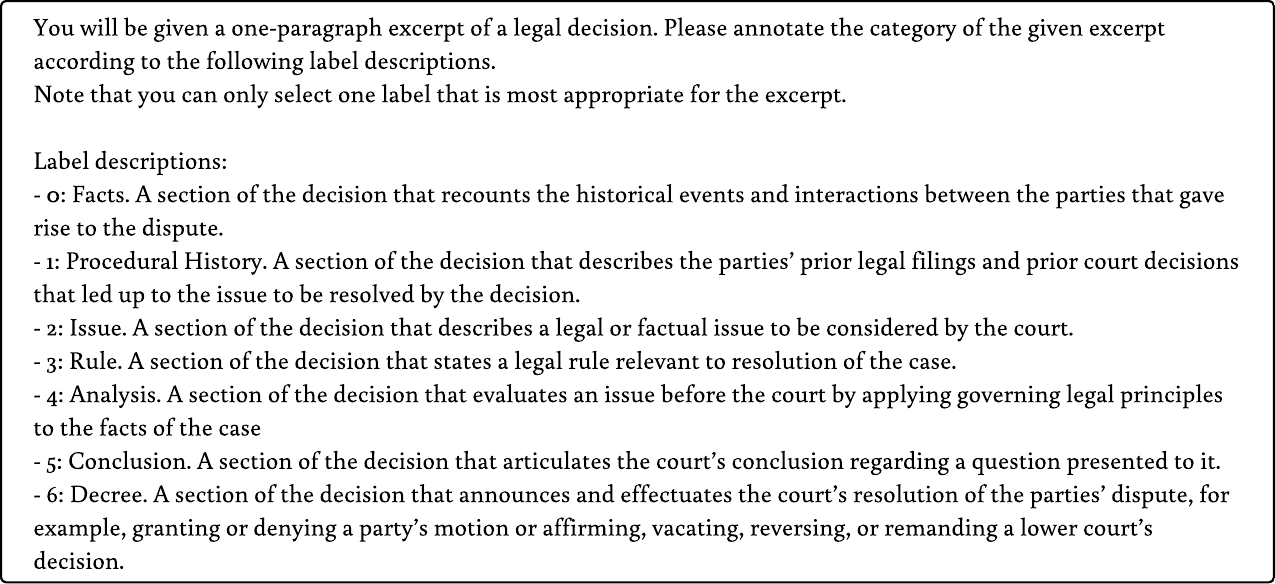}
  \caption{The annotation guideline of FoDS dataset.}
  \label{fig:guideline-fods}
\end{figure*}

We provide annotation guidelines of each dataset from Figure~\ref{fig:guideline-refind} to Figure~\ref{fig:guideline-fods}.

\subsection{Prompt Templates}\label{subsec:appendix-prompt}
\begin{figure*}[b]
  \centering
  \includegraphics[width=\linewidth]{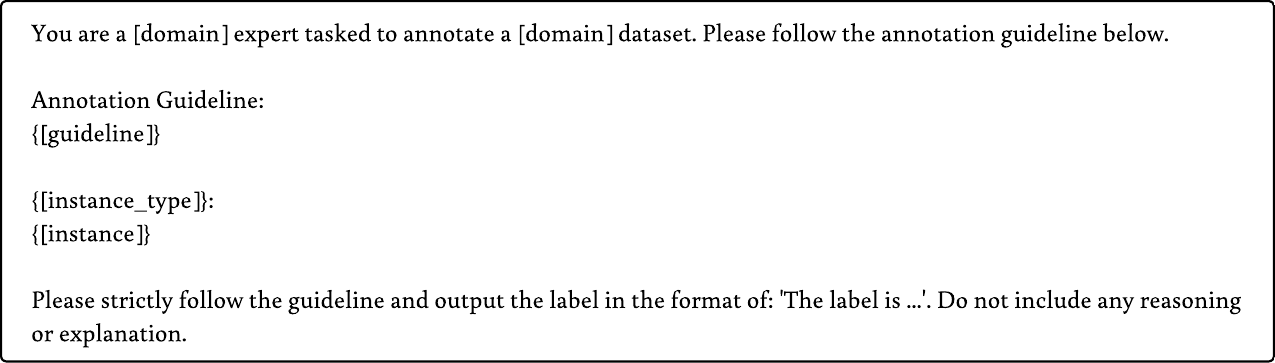}
  \caption{Vanilla prompt template.}
  \label{fig:template-vanilla}
\end{figure*}

\begin{figure*}[b]
  \centering
  \includegraphics[width=\linewidth]{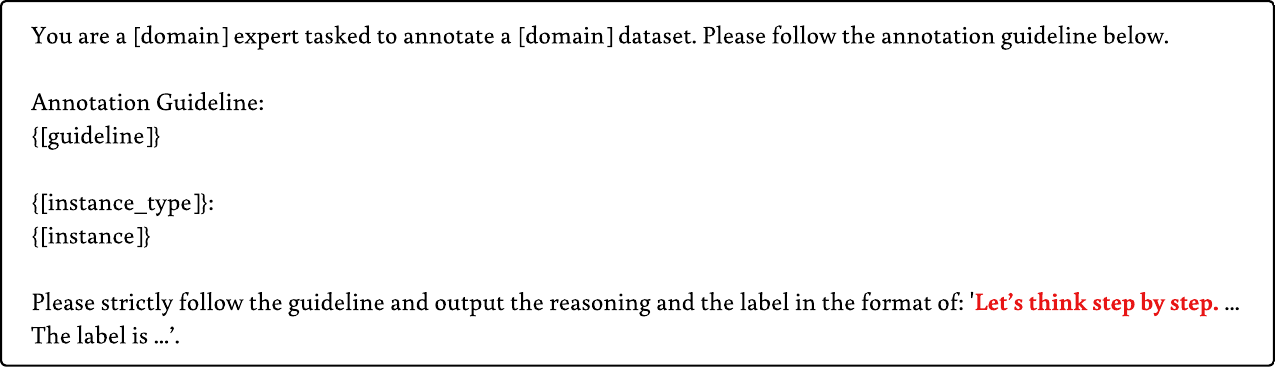}
  \caption{Chain-of-Thought prompt template.}
  \label{fig:template-cot}
\end{figure*}

\begin{figure*}[t]
  \centering
  \includegraphics[width=\linewidth]{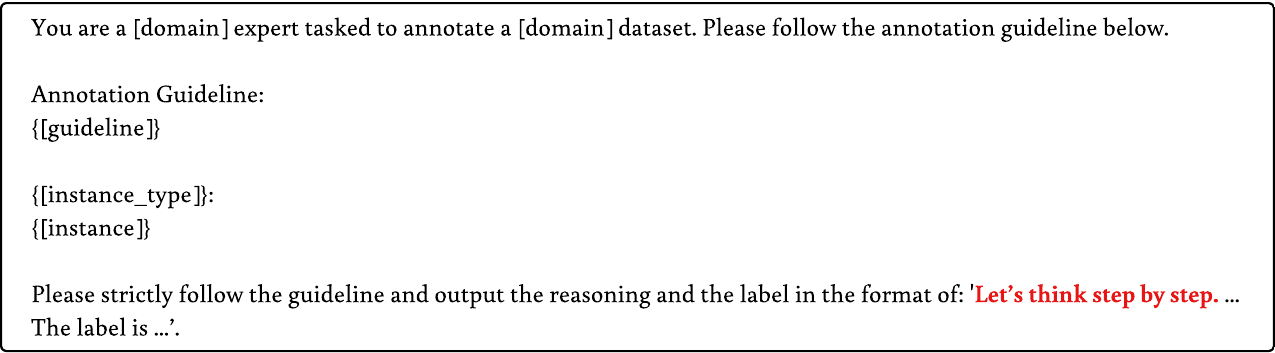}
  \caption{Self-Refine prompt template. Step 1: Generate.}
  \label{fig:template-self-refine-1}
\end{figure*}

\begin{figure*}[t]
  \centering
  \includegraphics[width=\linewidth]{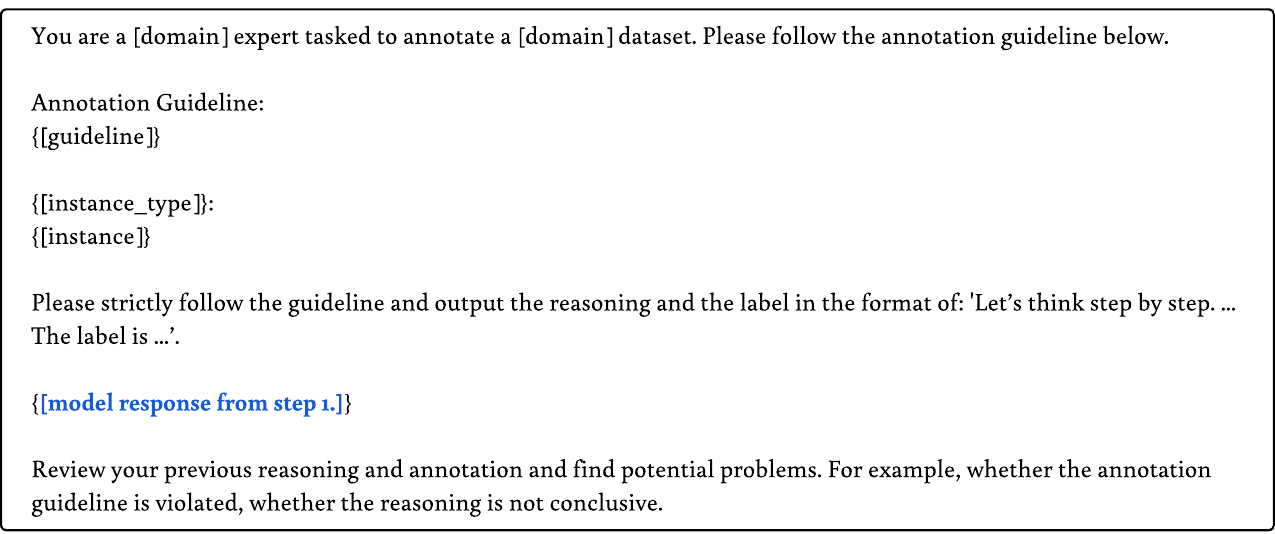}
  \caption{Self-Refine prompt template. Step 2: Review.}
  \label{fig:template-self-refine-2}
\end{figure*}

\begin{figure*}[t]
  \centering
  \includegraphics[width=\linewidth]{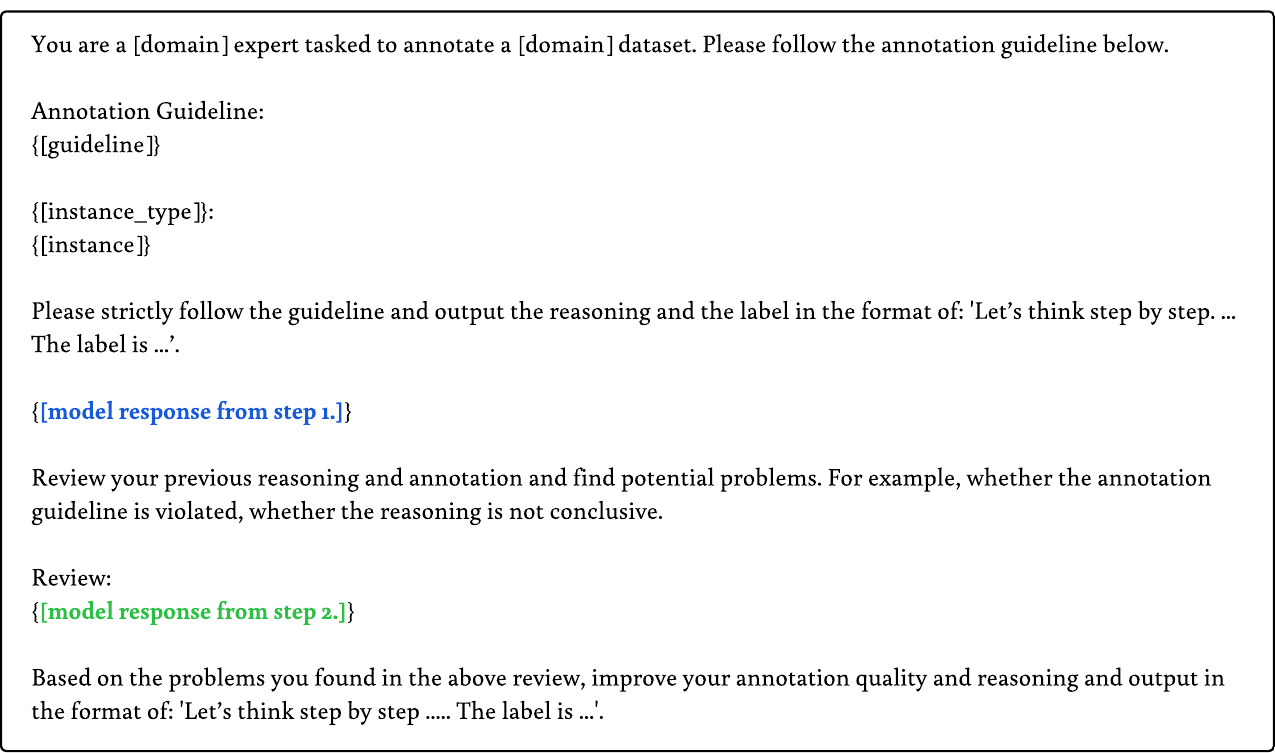}
  \caption{Self-Refine prompt template. Step 3: Refine.}
  \label{fig:template-self-refine-3}
\end{figure*}

\begin{figure*}[t]
  \centering
  \includegraphics[width=\linewidth]{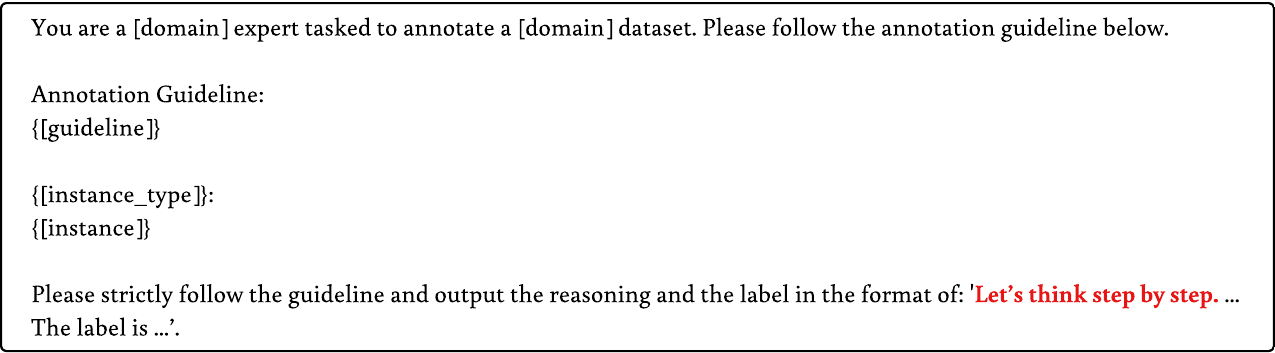}
  \caption{Multi-agent peer-discussion prompt template. Step 1: Generate initial annotation.}
  \label{fig:template-multi-agent-1}
\end{figure*}

\begin{figure*}[t]
  \centering
  \includegraphics[width=\linewidth]{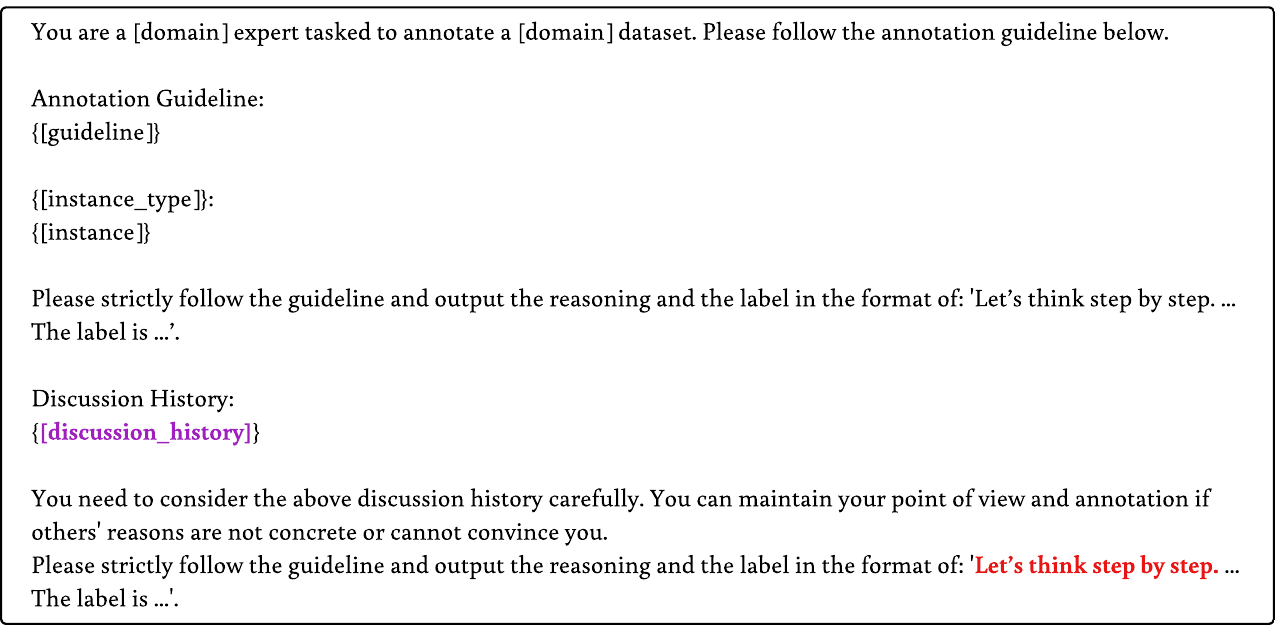}
  \caption{Multi-agent peer-discussion prompt template. Step 2: Discuss and re-annotate.}
  \label{fig:template-multi-agent-2}
\end{figure*}

We provide prompt templates of each methods from Figure~\ref{fig:template-vanilla} to Figure~\ref{fig:template-multi-agent-2}.

\begin{table*}[tp!]
  \small
  \centering
  \addtolength{\tabcolsep}{1pt}
  \resizebox{0.9\linewidth}{!}{
    \begin{tabular}{lllrr}
    \toprule
    \textbf{Domain} & \textbf{Dataset} & \textbf{Instance Type} & \textbf{$\#$Instances} & \textbf{$\#$Labels} \\
    \midrule
    \multirow{2}[2]{*}{Finance} & REFinD~\citep{kaur2023refind} & Sentence & 500 & 29 \\
          & FOMC~\citep{shah2023trillion}  & Sentence & 500 & 3 \\
    \midrule
    \multirow{2}[2]{*}{Biomedicine} & AP-Rel~\citep{gao2022hierarchical} & Pair  & 73 & 4 \\
          & CODA-19~\citep{huang2020coda} & Paper Abstract & 508 & 5 \\
    \midrule
    \multirow{2}[2]{*}{Law} & CUAD~\citep{hendrycks2021cuad}  & Clause & 500 & 32 \\
          & FoDS~\citep{guha2024legalbench}  & Excerpt & 367 & 7 \\
    \bottomrule
    \end{tabular}%
    }
    \caption{The statistics of existing domain-specific datasets.}
    \label{tab:dataset-statistic}
\end{table*}%

\end{document}